\documentclass[11pt,a4paper]{article}
\usepackage[hyperref]{acl2021}
\usepackage{times}
\usepackage{latexsym}

\usepackage[inline]{enumitem}
\usepackage{fancyvrb}
\usepackage{footnote}
\makesavenoteenv{tabular}

\usepackage[utf8]{inputenc}
\DeclareUnicodeCharacter{2665}{\O}

\usepackage{graphicx}
\usepackage{multirow}
\usepackage{booktabs}
\usepackage{comment}
\usepackage{tikz}
\usepackage{amsmath}
\usepackage{bm}
\usepackage{amssymb}
\usepackage{subcaption}
\usetikzlibrary{decorations.markings}

\aclfinalcopy 

\title{Detecting Bot-Generated Text by Characterizing Linguistic Accommodation in Human-Bot Interactions}

\author{Paras Bhatt \and Anthony Rios\\
 Department of Information Systems and Cyber Security \\
  University of Texas at San Antonio \\
  \texttt{\{Paras.Bhatt, Anthony.Rios\}@utsa.edu} \\}

\date{}

\begin{document}
\maketitle
\begin{abstract}

Language generation models' democratization benefits many domains, from answering health-related questions to enhancing education by providing AI-driven tutoring services. However, language generation models' democratization also makes it easier to generate human-like text at-scale for nefarious activities, from spreading misinformation to targeting specific groups with hate speech. Thus, it is essential to understand how people interact with bots and develop methods to detect bot-generated text. This paper shows that bot-generated text detection methods are more robust across datasets and models if we use information about how people respond to it rather than using the bot's text directly. We also analyze linguistic alignment, providing insight into differences between human-human and human-bot conversations.

\end{abstract}

\section{Introduction}

Bots are useful in a wide variety of applications areas including  business~\cite{kaczorowska2019chatbots}, education~\cite{kerlyl2006bringing}, and health~\cite{yadav2019feedpal,liednikova2020learning}. For instance, \citet{yadav2019feedpal} studied the use of chatbots as a drop-in first-point-of-contact for women in India seeking breastfeeding information. Similarly, researchers have studied bots to answer COVID-19-related questions~\cite{oniani2020qualitative} and screen individuals for risks of contracting the virus~\cite{martin2020artificial}. Overall, the wide availability of software packages, tools, and pre-trained models has democratized the creation of bots.


Even with the increasing interest in bots for social good (e.g., COVID-related chatbots), there is still a concern regarding their abuse to spread misinformation, be used for targeted discrimination, deceive users, and perform fraud~\cite{daniel2019bots}.
Given the potential good and harm bots can create, it is essential to study how the bots \textit{should} act and how people \textit{do} interact with such bots for specific~applications.

Automatically identifying bots 
online is well studied~\cite{garcia2019empirical,herzig2019bot2vec,kosmajac2019twitter,ippolito2020automatic,jawahar2020automatic}. Bot detection methods rely on two forms of information~\cite{orabi2020detection}: behavior and content. Behavior relates to measuring how often bots post, the time posts are created, and conversational network structures~\cite{beskow2018bot}. Content involves using the bot's text directly. Focusing on text suffers from generalization issues, making it challenging to detect bots that discuss different topics or operate in different domains. Likewise, behavioral approaches assume that bots will behave differently than people at a superficial level, such as posting more often than humans. Even in the network analysis of conversation structures~\cite{beskow2018bot}, many of the human-bot interactions on social media are not human-like interactions. For instance, it is understandable that humans will interact with a bot that converts pounds to kilograms differently than other humans. Given bots' current use-cases ranging from counselors to healthcare information providers, it is vital to understand how humans and bots interact beyond trivial~applications.

Before looking at how humans interact with bots, it is essential to understand how humans interact. This paper focuses on Communication Accommodation Theory (CAT) and general language use to analyze interactions. CAT is used to study language use in various domains to understand human behavior~\cite{giles1973towards,tausczik2010psychological}. More specifically, we study linguistic accommodation---where speakers come to talk more (or less) similarly as they interact---which has been analyzed in mental health support on Reddit, showing a positive link with informational and emotional support~\cite{sharma2018mental}. The alignment of two people in terms of linguistic style has positively predicted successful outcomes of negotiations~\cite{taylor2008linguistic}, and doctors are recommended to accommodate perspective to improve patient~care~\cite{kline1984person,wood2019departing}. The way people use and accommodate pronouns can indicate power, where high-status individuals use ``I'' less and ``you/we'' words more than low-status individuals~\cite{kacewicz2014pronoun}.

Linguistic accommodation in human-bot interactions has been studied in system design, showcasing that there is a strong link between user experience and language style of the system~\cite{chaves2019s,chaves2020should,thomas2020expressions}. Recent research has also shown a link between language style in chatbots and user engagement for e-commerce~\cite{elsholz2019exploring}.
Our focus is to understand real long open-domain human-bot interactions better. Our study can also influence how bots are detected on social media and provide a better understanding of how human-human interactions differ from human-bot interactions is essential for bot development. For instance, when should bots accommodate towards the user? Can we tell how well a system performs for a specific task based on whether the user accommodates the bot? If human-bot conversations are linguistically different than human-human interactions, research from human-human studies may not generalize to human-bot interactions.

Toward addressing the potential societal impacts of open-domain bots, this paper addresses the following research questions (RQs) in the context of bot Detection:
\begin{enumerate}[label=\textbf{RQ\arabic*.},start=1,leftmargin=*]
    \item How do humans and bots align in human-bot interactions?
    \item How does the alignment in human-bot interactions compare to the alignment in human-human interactions?
    \item Are differences in alignment between human-human and human-bot interactions similar across domains and language generation learning methods?
\end{enumerate} 
 Overall, we analyze whether human users' language changes stylistically with bots as compared to human-human conversations. Furthermore, we show that small differences in language style provides robust information, compared to the bots' language patterns, to accurately detect bots.

\section{Related Work}


\vspace{3mm} \noindent \textbf{Bot Detection.} Detecting bots in the wild is a widely studied problem~\cite{garcia2019empirical,herzig2019bot2vec,kosmajac2019twitter,ippolito2020automatic,jawahar2020automatic}. bot detection methods rely on two forms of information~\cite{orabi2020detection}: Content and Behavior. \citet{garcia2019empirical} studied how pre-trained language models perform for the task of bot detection. \citet{kosmajac2019twitter} developed ``language-independent'' stylistic features that measure language diversity to detect bots. \citet{knauth2019language} empirically explored content and behavioral features for bot detection. \citet{beskow2018bot} analyzed human-bot interactions by developing methods to detect bots using network analysis of conversation structures. Much of the prior work has focused on detecting bots ``in the wild.`` Thus, many of the interactions between bots and humans are superficial (e.g., receiving movie quotes from a bot). This paper differs from prior work in two ways. First, rather than using content from the bot or general behavioral information (e.g., post frequency), we try to detect bots by analyzing how humans respond to them (compared to human-human interactions). Second, rather than exploring bots in the wild, we explore two types of datasets: one where researchers instruct participants to interact with bots and humans in the same way, and another dataset where researchers instruct participants to converse with a bot.

\vspace{3mm} \noindent \textbf{Analyzing Human-Human Conversations.}
Linguistic alignment is known to interact with a wide array of social factors. For instance, the level of alignment of people's linguistic style in a conversation has been claimed to be affected by their relative social power~\cite{gnisci2005sequential,xu2018not,danescu2011mark}. Cooperative decision-making tasks are positively related to the participants' linguistic convergence~\cite{fusaroli2012coming,kacewicz2014pronoun}. Recently, \citet{sharma2018mental} analyzed mental health support forums on Reddit, showing that linguistic accommodation is positively linked with informational and emotional support.  Similarly, \citet{taylor2008linguistic} analyzed negotiation outcomes and linguistic alignment, showing that alignment positively predicted successful negotiation outcomes. There is also the support of linguistic alignment by doctors to enhance patient~care by improving trust and adherence to a treatment plan~\cite{kline1984person,wood2019departing}. Alignment is a powerful tool that can be used to improve understanding, trust, and potentially patient outcomes. Therefore, understanding \textit{when} and \textit{how} people should align with each other is an important area of research. Furthermore, it is essential to draw a line between engaging with a culture and language style and appropriating it, thereby potentially causing harm rather than building rapport with the partners in a conversation.

With the goal of understanding human-human interactions, there has also been progress in linguistic alignment measurement techniques~\cite{niederhoffer2002linguistic,danescu2011mark,jones2014finding,wang2014linguistic,doyle2016investigating,shin2018alignment}. \citet{danescu2011mark} presented an easy-to-compute expression that measures the increase in the conditional probability given that a conversational partner has used it. An issue with the method proposed by \citet{danescu2011mark} is that it assumes messages between two people have similar length. To overcome this limitation, \citet{doyle2016investigating} introduced the Word-Based Hierarchical Alignment Model (WHAM), a hierarchical graphical model where the parameters are learned using Bayesian inference. Another method called the Simplified Word-Based Alignment Model (SWAM) was recently proposed by \citet{shin2018alignment}. SWAM attempts to compare alignment between different groups when the alignment scores are assumed to differ substantially. Unfortunately, SWAM only estimates group-level alignment, not conversation-level between two specific users/bots.
Linguistic alignment has also been shown to be predictive of specific tasks. In this paper, we use the method proposed by \citet{danescu2011mark} to estimate linguistic alignment. For instance, \citet{niven2019detecting} use alignment features to predict discourse acts.



\vspace{3mm} \noindent \textbf{Analyzing Human-Bot Conversations.}
Research studying the interaction between bots and humans has been explored from a wide array of perspectives. For example, systems that use emotionally expressive interjections (``wow'', ``ahem'') in their text to speech responses can significantly improve the user experience~\cite{cohn2019large}. Given the popularity of bots in application areas from business~\cite{kaczorowska2019chatbots} to healthcare~\cite{pieraccini2009we}, it is also important to understand how language generation style and alignment impacts their intended use. 
There has been a recent interest in analyzing accommodation and similar concepts in human-bot interactions. For instance, \citet{ahn2020code} show that humans will match \textit{code-switching} patterns introduced by a chat system. Moreover, users tend to have positive reactions towards systems that code-switch. \citet{ma2020empirical} measure \textit{lexical entrainment} between a specific Reddit bot and users. They find that sentiment of  bot has a positive effect on the sentiment of the humans response. Furthermore, \citet{ma2020empirical} show that human responses tend to overlap with the bots original post. Compared to prior work studying accommodation-related aspects of human-bot interactions, this paper differs in two ways. First, just analyzing linguistic alignment, we ground our study of human-bot interactions in the real-world task of bot detection. Second, we analyze cross-domain generalization of accommodation patterns in human-bot interactions, where cross-domain includes types of bots (e.g., retrieval and transformer-based models), data collection procedures (e.g., sampling bias), and conversation topics.

\section{Datasets}

\begin{figure}[t]
\centering
\includegraphics[width=.65\linewidth]{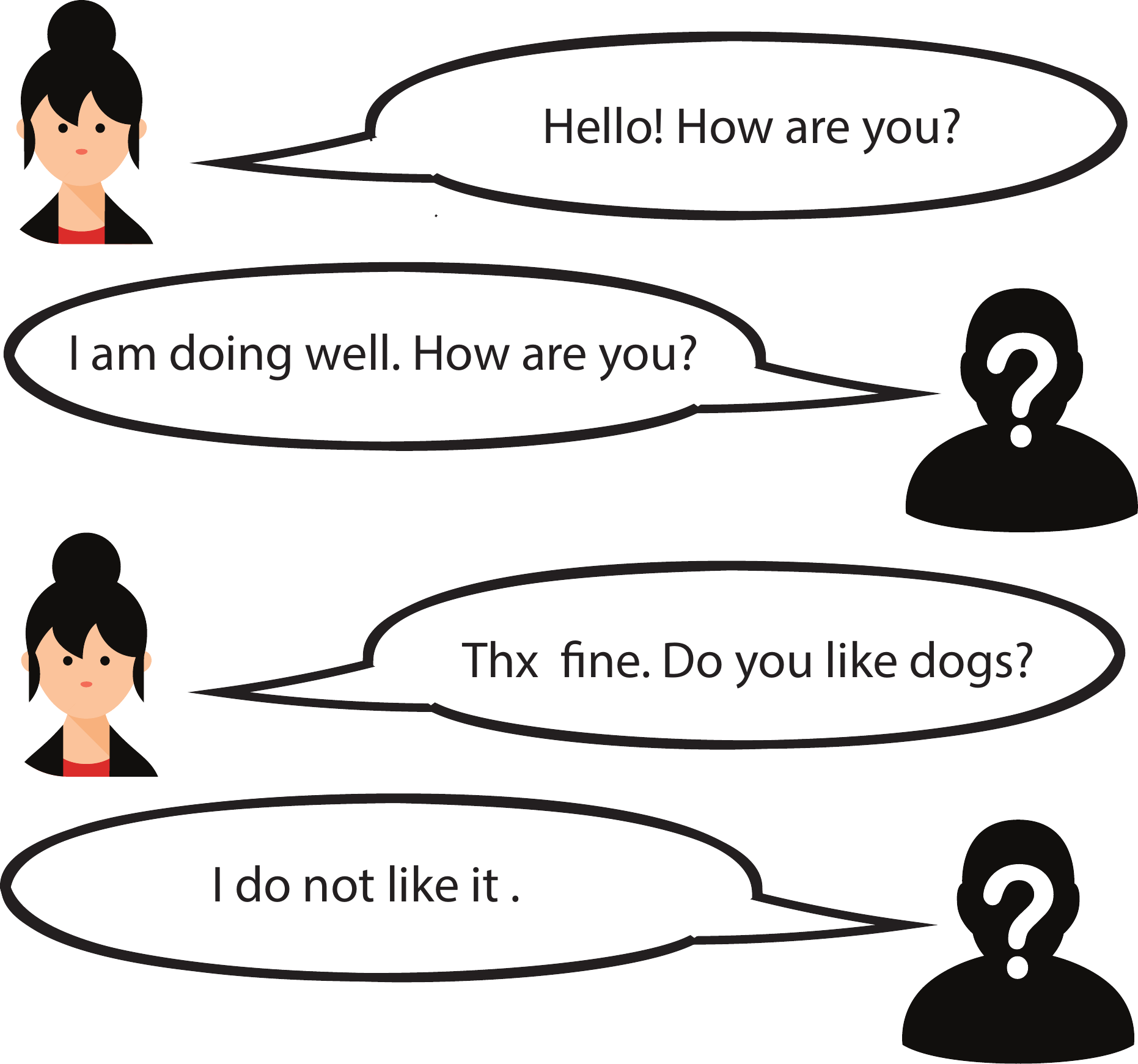}
\caption{This figure depicts an potential conversation between a known human user and an unknown user. The unknown user may be a human or a bot.}
\label{fig:overview}
\end{figure}

An overview of the task we introduce in this paper is shown in Figure~\ref{fig:overview}. We formulate a bot detection task between two entities, either a human and bot or a human and human. We assume that one entity is always human, and the other entity is unknown---either a human or a bot. Formally, let $D = [u^h_1, u^o_2, \ldots , u^h_{N-1}, u^o_N]$ represents a sequence of utterances, where $u^h_i$ represents the \textit{i}-th utterance in a conversation and that it was made by a human $h$. Likewise, $u^o_i$ represents an unknown entity's utterance (human or bot). $N$ is the total number of utterances in the conversation. Our goal is to develop a classifier $f(D)$ that maps to a class in the set $T = \{\text{human-human}, \text{human-bot}\}$, where the human-bot means a bot is a part of the conversation. We formulate two bot detection datasets for this task consisting of three known bot datasets: ConvAI2, WOCHAT, and DailyDialog. The basic statistics of each dataset are shown in Table~\ref{tab:DataStats}. WOCHAT and DailyDialog are used to form a Control dataset. We describe each dataset~below:

\begin{table}[t]
\resizebox{\linewidth}{!}{%
\begin{tabular}{lrrrr}
\toprule
Dataset   Name & \# $D$ & \# $u$ & \shortstack{Avg. $u$} & \shortstack{Avg. W} \\ \midrule
\multicolumn{5}{c}{ConvAI2 Data} \\ \midrule
INTERMEDIATE   & 291 & 4317 & 14.83 & 6.66 \\
TOLOKERS       & 3127 & 39155 & 12.52 & 7.05 \\
VOLUNTEER       & 1111 & 14623 & 13.16 & 6.60 \\
PERSONA-CHAT   & 18878 & 278478 & 14.75 & 1.85 \\ \midrule \midrule
\multicolumn{5}{c}{Control Data} \\ \midrule
IRIS           & 163 & 5687 & 34.89 & 5.46 \\
TICKTOCK       & 206 & 5462 & 26.51 & 7.07 \\
DailyDialog       & 13118 & 102980   & 7.85 & 13.08 \\ \bottomrule
\end{tabular}}
\caption{Summary of each dataset, including the number of dialogues (\# $D$), average number of utterances per dialogue (Avg. $u$), average number of words per utterance (Avg. W) and total number of utterances~(U)}
\label{tab:DataStats}
\end{table}

\vspace{3mm} \noindent \textbf{ConvAI2.} We use four datasets from the Second Conversational Intelligence Challenge (ConvAI2). The motivation behind the ConvAI2 challenge was to develop new approaches towards open-domain chatbots~\cite{zhang2018personalizing,dinan2019second}. Overall, we use two types of ConvAI2 datasets: training datasets containing human-human interactions and evaluation phase datasets containing human-bot interactions. First, we use the PERSONA-CHAT dataset, a collection of human-human interactions where researchers instructed Mechanical Turk users to converse with one another, assuming specific personas (i.e., profile descriptions). The ConvAI2 competition used automatic (e.g., Perplexity) and human evaluation procedures. The human evaluation procedures involve either paid workers or volunteers that interact with models built using PERSONA-CHAT, after which the volunteers/workers are asked to rate their interactions. Moreover, the human evaluator is instructed to interact with the bots in the same way as the Mechanical Turk users who participated in creating the PERSONA-CHAT dataset. 

We use three ``evaluation-phase'' datasets~\footnote{\url{http://convai.io/data/}}: TOLOKERS, VOLUNTEERS, and INTERMEDIATE. The TOLOKERS used solicited workers to chat with the models, similar to the PERSONA-CHAT's Mechanical Turk setup. Specifically, TOLOKERS consists of data collected during DeepHack.Chat~\footnote{\url{http://deephack.me/chat}} hackathon via paid workers using the Yandex.Toloka service. The INTERMEDIATE dataset consists of more dialogues by the bots from DeepHack.Chat, but the interactions come from volunteers. Finally, the VOLUNTEER dataset was collected during the final ``wild evaluation'' round of the ConvAI2 competition. Human-bot interactions were collected from volunteers through the Facebook Messenger and Telegram APIs.








\vspace{3mm} \noindent \textbf{Control Dataset.}
One of the research questions we explore in this paper is related to out-of-domain performance. Specifically, can we detect bots based on human response when data were collected under different settings and where the conversation topics differ?
To address the out-of-domain research question, we use two additional datasets collected in the The Workshop on Chatbots and Conversational Agent Technologies (WOCHAT)~\cite{kong2019subjective}. As part of a shared task, the workshop makes several bots available, has participants contribute new bots, and participants interact with the bots providing utterance-level feedback regarding their performance. We use two human-bot interaction datasets released by the organizers: IRIS and the TickTock dataset.~\footnote{\url{http://workshop.colips.org/wochat/data/index.html}}
both IRIS~\cite{banchs2012iris} and TickTock~\cite{yu2015ticktock} are  retrieval-based bot variants that were ``trained'' on different datasets. Unlike the ConvAI2 task, humans are not instructed to converse with the bots as if they are another human. Thus, sometimes humans will ask things such as, ``Are you a Robot?''. This point provides a unique aspect that increases cross-data differences.

Unfortunately, human-human conversations were not released from the WOCHAT shared task. The focus was annotating human-bot interactions. Thus, we need to augment WOCHAT with real human-human conversations. Our study uses the DailyDialog dataset as our source of human-human interactions~\cite{li2017dailydialog}. Unlike the ConvAI2-related data, DailyDialog consists of conversations between English learners practicing everyday English dialog in daily life.

\vspace{3mm} \noindent \textbf{Data Processing.} We split the data into three groups: Unpaid (U), Paid (P), and Control (C). The U dataset consists of PERSONA-CHAT, INTERMEDIATE, and VOLUNTEER ConvAI2 datasets. The P dataset consists of PERSONA-CHAT and TOLOKERS---the name Paid comes from the fact that the TOLOKERS were paid as part of a crowdsourcing task. We split this group from the others in case this affects conversation behavior. The C dataset consists of IRIS, TICKTOCK, and DailyDialog. Each dataset is divided into 70\%, 10\%, and 20\% training, validation, and test splits, respectively. It is important to note that the same PERSONA-CHAT training, validation, and test examples are used in both the U and P datasets.

\section{Method}\label{sec:methods}

To detect human-bot conversations, we explore two types of features below: Content and Stylistic features.

\subsection{Content Features}

We define content information as features describing ``\textit{what}'' humans and bots say in their interactions. Specifically, we describe two sets of content features: bag-of-words and embeddings.

\vspace{3mm} \noindent \textbf{Bag-of-words.} As a simple baseline, we use TF-IDF-weighted unigrams from a dialog to detect whether a bot is part of the conversation. We explore three settings for the TF-IDF features: human-only, unknown-only, and human-unknown. The \textit{unknown} user in the conversation can be either a bot or a human. Thus, in the human-Only setting, only the human's unigrams from each dialogue $D_{h} = [u^h_1, u^h_3, \ldots, u^h_{N-1}]$ are used to detect human-bot interactions. The unknown-only setting uses only the unknown user's unigrams from each dialogue $D_{o} = [u^o_2, u^o_4, \ldots, u^o_{N}]$, and the human-unknown setting uses both the human's and unknown user's unigrams from each dialogue $D = [u^h_1, u^o_2, \ldots, u^h_{N-1}, u^o_{N}]$.

\vspace{3mm} \noindent \textbf{Embedding Features.} We encoded each dialogue $D$ using BERT~\cite{devlin2019bert}\footnote{We use the bert-base-uncased pre-trained model available in the HuggingFace package~\cite{wolf2019huggingface}.} by feeding the first 510 WordPieces and then averaging the word representations extracted from the second-to-last layer. We explore three variants, human-Only, Unknown-Only, and human-Unknown. Depending on the variant, the exact WordPieces passed to BERT change. For instance, in the human-Only setting, the first 510 WordPieces made by the human from a concatenation of all of the humans utterances $D_{h} = [u^h_1, u^h_3, \ldots, u^h_{N-1}]$ are passed to BERT. Similarly, all of the unknown user's utterances are used for the unknown-only variant to obtain $D_{o} = [u^o_2, u^o_4, \ldots, u^o_{N}]$. In the human-unknown setting, all human and unknown utterances are concatenated in $D = [u^h_1, u^o_2, \ldots, u^h_{N-1}, u^o_{N}]$ before generating features with BERT.

\subsection{Stylistic Features}

The stylistic features encode ``\textit{how}'' humans and bots speak in their interactions. We use two sets of stylistic features: Linguistic Inquiry and Word Count (LIWC)~\cite{pennebaker2015development} and linguistic accommodation~\cite{danescu2011mark}.

\vspace{3mm} \noindent \textbf{LIWC.} We experiment with the psychologically validated word categories (e.g., positive emotion, cognitive, and social processes) in LIWC as features. These lexicons might reveal more about a writer’s thought processes, emotional states, and intentions. For LIWC features, we use the same word categories described in the Linguistic Accommodation Section below to train our LIWC-based classifier. Specifically, we use LIWC to process each utterance. Next, each utterance's LIWC scores are averaged together to form a dialogue-specific feature vector. We experiment with LIWC features variants: human-only, unknown-only, and human-unknown. Each setting averages the LIWC scores across a different set of utterances, similar to the bag-of-words and embedding~features.

\vspace{3mm} \noindent \textbf{Linguistic Accommodation.} We use the method proposed by \citet{danescu2011mark} to analyze linguistic alignment. It includes two primary group-level and conversation-level measures: baseline word usage and alignment. The group-level baseline word use corresponds to the rate at which a person uses a given \textit{word category} when it has not been used previously in a conversation. The group-level alignment score reflects the proportionate increase---compared to the baseline score---in the likelihood of the word being used when it has been used previously in the conversation. Similarly, the conversation-level metrics measure accommodation and usage between a human and Unknown (bot or human) user in the each dialogue $D_i$. We measure the probability of seeing a word category (e.g., LIWC categories) $w_c$ given $w_c$ appeared in the previous utterance $P(w_c \in u^o_i| w_c \in u^h_{i-1})$\footnote{The probabilities are from the perspective of the Unknown user in the conversation. They are also calculated from the human's perspective.} and the baseline probability of seeing $w_c$ in the conversation $P(w_c \in u^o_i)$. Next, the accommodation score is calculated by taking the difference between both probabilities, $acc(c) = P(w_c \in u^o_i| w_c \in u^h_{i-1}) - P(w_c \in u^o_i)$. These empirical probabilities are at the conversation level (i.e., measuring how two specific users align). To obtain the group-level estimates, we simply average the accommodation scores $acc(c)$ over all conversations $D_i$ for each word category.
We use the following 17 LIWC categories~\cite{pennebaker2015development}: i, you, we, they, social, cogproc, posemo, negemo, article, prep, certain, conj, discrep, negate, pronoun, quant, and tentat.
We experiment with two main settings as described for the previous methods: Human-Only, Unknown-Only. The Human-Only setting consists of just using the known human's alignment scores in each conversation and vice-versa for the Unknown-Only~setting. 

\vspace{3mm} \noindent \textbf{Model Training Details.} For the content feature sets, we train a Logistic Regression classifier from the Scikit-Learn package~\cite{pedregosa2011scikit}. Using the validation split for each dataset, we grid-search over the C-values \{.0001, .001, .01, .1,~1.,~10.\}, the logistic regression class weight parameters \{None,~balanced\}, and normalization procedures \{standardize, unit normalize, None\}. For the Stylistic features, we train a Random Forest classifier from  the Scikit-Learn package. Again, using the validation split, we grid-search over the class weight parameters \{None, balanced, balanced\_subsample\}, criterion measures \{gini, entropy\}, max features \{sqrt, log2, None\}, and  bootstrap parameters \{True, False\}. For all experiments using the Random Forest classifier, we set n\_estimators to 1000.

\begin{table}[t]
\centering
\resizebox{\linewidth}{!}{%
\begin{tabular}{@{}llll@{}}
\toprule
                         & U $\to$ U & P $\to$ P & C $\to$ C \\ \midrule \midrule
                        \multicolumn{4}{c}{Baselines} \\ \midrule
Most Frequent            &       .482      &      .462       &      .493       \\
Most Infrequent          &     .065        &      .125       &       .027      \\
Stratified (random)      &     .491        &        .498     &     .487        \\ \midrule \midrule
                        \multicolumn{4}{c}{Content Features} \\ \midrule
Human Bag-of-Words       &       .980      &      .987       &      .939       \\
Human BERT               &       .989      &     .996       &     .987        \\ \cmidrule{2-4}
Unknown Bag-of-Words         &     .971        &       .970      &       .997      \\
Unknown BERT                 &       .996      &       .996      &       .983      \\  \cmidrule{2-4}
Human + Unknown Bag-of-Words &       .958      &        .963     &         .986    \\
Human + Unknown BERT         &    .990        &     .992        &      .990       \\ \midrule  \midrule
                        \multicolumn{4}{c}{Stylistic Features} \\ \midrule
Human LIWC               &       .878      &     .899        &      .705       \\
Human Accommodation      &       .989      &       .987      &      .627       \\  \cmidrule{2-4}
Unknown LIWC                 &     .838        &     .862        &     .707        \\
Unknown Accommodation        &       .887      &      .897       &      .674     \\   \cmidrule{2-4}
Human \& Unknown LIWC        &      .885       &     .903        &  .601 \\ 
Human \& Unknown  Accommodation        &     .988        &          .990   &     .729        \\ 
\bottomrule
\end{tabular}%
}
\caption{Source $\to$ Source Macro F1 Results for Bot Detection}
\label{tab:res}
\end{table}

\begin{table*}[t]
\centering
\resizebox{.9\textwidth}{!}{
\begin{tabular}{lllllll}
\toprule
 & &
  \textbf{U $\to$ C} &
  \textbf{P $\to$ C} &
  \textbf{C $\to$ U} &
  \textbf{C $\to$ P} &
  \textbf{AVG} \\ \midrule
\multirow{3}{*}{\textit{\shortstack[l]{\text{Baselines}}}} & 1. Most Frequent                  &  .493 & .493 & .482 & .462 &  .483 \\
 &  2. Most Infrequent              & .027 & .027 &  .065 & .125 & .061 \\
 &  3. Stratified (Random)         & .502 & .479 & .499 & .476 & .489 \\ \midrule
\multirow{6}{*}{\shortstack[l]{\textit{Content:}\\\textit{What the bots and}\\\textit{humans write?}}} &  4. Human bag-of-words  &  .518 & .504 & .536 & .608 &  .541   \\
 & 5. Human BERT             & .493 & .493 & .482 & .462 & .483 \\ \cmidrule(lr){2-7}
 &  6. Unknown bag-of-words              & .493 & .493 & .535 & .475 &  .499   \\
 &  7. Unknown BERT                 & .493 & .493 & .482  & .462 & .483 \\ \cmidrule(lr){2-7}
 &  8. Unknown and Human bag-of-words            & .509 & .510 & .639 & .608 & .567 \\
 & 9. Unknown and Human BERT                 & .493 & .493 & .522 & .478 & .497 \\
 \midrule                        
\multirow{9}{*}{\shortstack[l]{\textit{Stylistic:}\\ \textit{How the bots and}\\\textit{humans write?}}} & 10. Human LIWC        & .480 & .491 & .524 & .512 & .502 \\
 &  11. Human Accommodation      & \textbf{.631} & .591 & .604 & .510 & .584 \\ 
  &  12. Human LIWC + Accommodation      & .605 & {.608} & \textbf{.703} & \textbf{.724} & \textbf{.660} \\ \cmidrule(lr){2-7}
 &  13. Unknown LIWC             & .474 & .478  & .483 & .476 & .478 \\
 &  14. Unknown Accommodation        & .428 & .424 & .503 & .494 & .462 \\
  &  15. Unknown LIWC + Accommodation        & .462 & .436 & .502 & .497  & .474 \\  \cmidrule(lr){2-7}
   &  13. Human \& Unknown LIWC             & .521 & .504   & .532 & .556 & .528 \\
 &  14. Human \& Unknown Accommodation        & .611 & .620  & .642  & .677 & .637 \\
  &  15. Human \& Unknown LIWC + Accommodation        & .622 & \textbf{.633} &.642  & .677  & .643  \\ \bottomrule
\end{tabular}
}
\caption{This table reports the cross-dataset Macro F1 score for detecting human-bot conversations for three datasets: Unpaid (U), Paid (P), and Control (C). The largest Macro F1 score in each column is in \textbf{bold}.\vspace{-1em}}
\label{tab:results}
\end{table*}

\section{Results}

In this section, we report two sets of results. First, in Subsection~\ref{sec:res:botdetect} we present the performance of the bot detection models we explain in Section~\ref{sec:methods}. Beyond the methods described in Section~\ref{sec:methods}, we also compare three baselines: Most Frequent, Most Infrequent, and Stratified. The Most Frequent baseline predicts the most frequent class for every example (i.e., human-human). The Most Infrequent baseline predicts the most infrequent class (i.e., human-bot) and the Stratified baseline makes random prediction proportional to each class's frequency. Second, in Subsection~\ref{sec:results:analysis}, we present a fine-grained analysis of the linguistic accommodation results on the ConvAI2 datasets. For all results, we report the Macro F1 (average F1 for the human-human and human-bot conversation classes).

\subsection{Bot Detection Experiments}\label{sec:res:botdetect}

The source dataset results are reported in Table~\ref{tab:res}. Overall, we find that content features are the most predictive for bot detection, when training and testing on the same train-test splits from the same dataset. Furthermore, we find that the BERT-based models are able to outperform the Bag-of-Words models on average. For instance, the Human Bag-of-Words model on dataset C obtains an F1 of .939. Yet, the Human BERT model obtains an F1 of .987. Likewise, for the stylistic features, we find that simply using LIWC works better than using the accommodation features alone for dataset C. However, combining both Accommodation and LIWC features from both the Human and the Bot is better than using either feature set individually with an F1 of .729.

The cross-dataset bot detection results are presented in Table~\ref{tab:results}. Specifically, the scores are from experiments where we train on a source dataset and evaluate each model on a target dataset's test split (i.e., source~$\to$~target). Overall, we make three major findings. First, for both Content and Stylistic features, we find that analyzing the known human's language in each conversation is more informative than analyzing the bot's content. For instance, the average (AVG) score for human bag-of-words is .541, while the bot bag-of-words model AVG result is nearly 4\% lower (.499). We have similar findings between human LIWC (.502) and bot LIWC (.478).  We find that combining both bot and human LIWC improves the AVG performance of the human-only LIWC model with an F1 of 0.528. Second, while the BERT-based model performs better when applied to data from the same datasets in Table~\ref{tab:res} (e.g., U~$\to$~U) in many settings, when the test dataset changes substantially (e.g., U~$\to$~C), the generalization performance of BERT drops compared to using bag-of-words. This result is potentially caused by overfitting to random source-specific characteristics. Third, we find that Accommodation features outperform all other individual feature sets with an AVG bot-detection Macro F1 of .584, with the exception of using both the human's and bot's accommodation features which has a Macro F1 of .637. Interestingly, human accommodation information (.584) is more predictive than the bot's text (.462). Intuitively, the type of responses generated by a bot can differ substantially depending on the bot's training data (e.g., ConvAI2 vs. DailyDialog) and model (e.g., transformer vs. retrieval-based models). Hence, the human's responses are more consistent with regard to alignment. The best combination overall is the combination of the  human's LIWC features with the human's accommodation features achieving a Macro F1 of .660. Yet, with the bot (Unknown) feature combination, the performance drops from .768 with Unknown LIWC to .674 after combining accommodation features.

\begin{table}[t]
\centering
\resizebox{0.65\linewidth}{!}{%
\begin{tabular}{lll}
\toprule
\textbf{Feature Group} & \textbf{Feature} & \textbf{FI} \\ \midrule
     Accomodation & pronoun & .598 \\
LIWC & social & .063 \\
LIWC & article & .043 \\
LIWC & prep & .028 \\
LIWC & pronoun & .020 \\
Accomodation & quant & .018 \\
LIWC & tentat & .017 \\
Accomodation & i & .017 \\
LIWC & you & .016 \\
LIWC & i & .015 \\  \bottomrule
\end{tabular}%
}
\caption{The top ten most \textit{informative} features for the Random Forest model trained on dataset P's human responses to detect human-bot conversations with LIWC and Accommodation features.\vspace{-1em}}
\label{tab:features}
\end{table}

\begin{figure*}[t]
        \centering
        \begin{subfigure}[b]{0.475\textwidth}
            \centering
            \includegraphics[width=\textwidth]{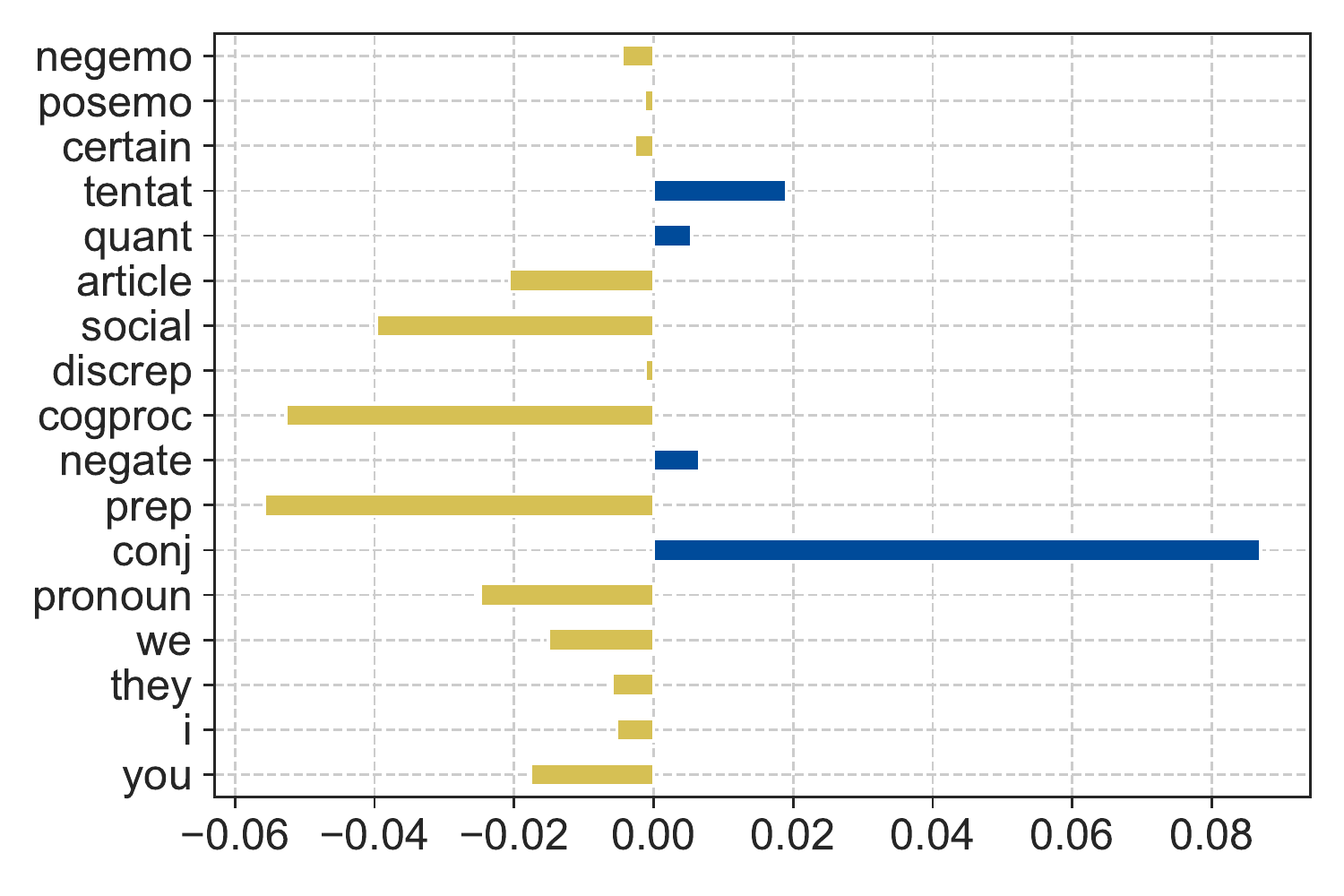}
            \caption[Network2]%
            {{ Human TickTock Alignment}}    
            \label{fig:human-ticktock}
        \end{subfigure}
        \hfill
        \begin{subfigure}[b]{0.475\textwidth}  
            \centering 
            \includegraphics[width=\textwidth]{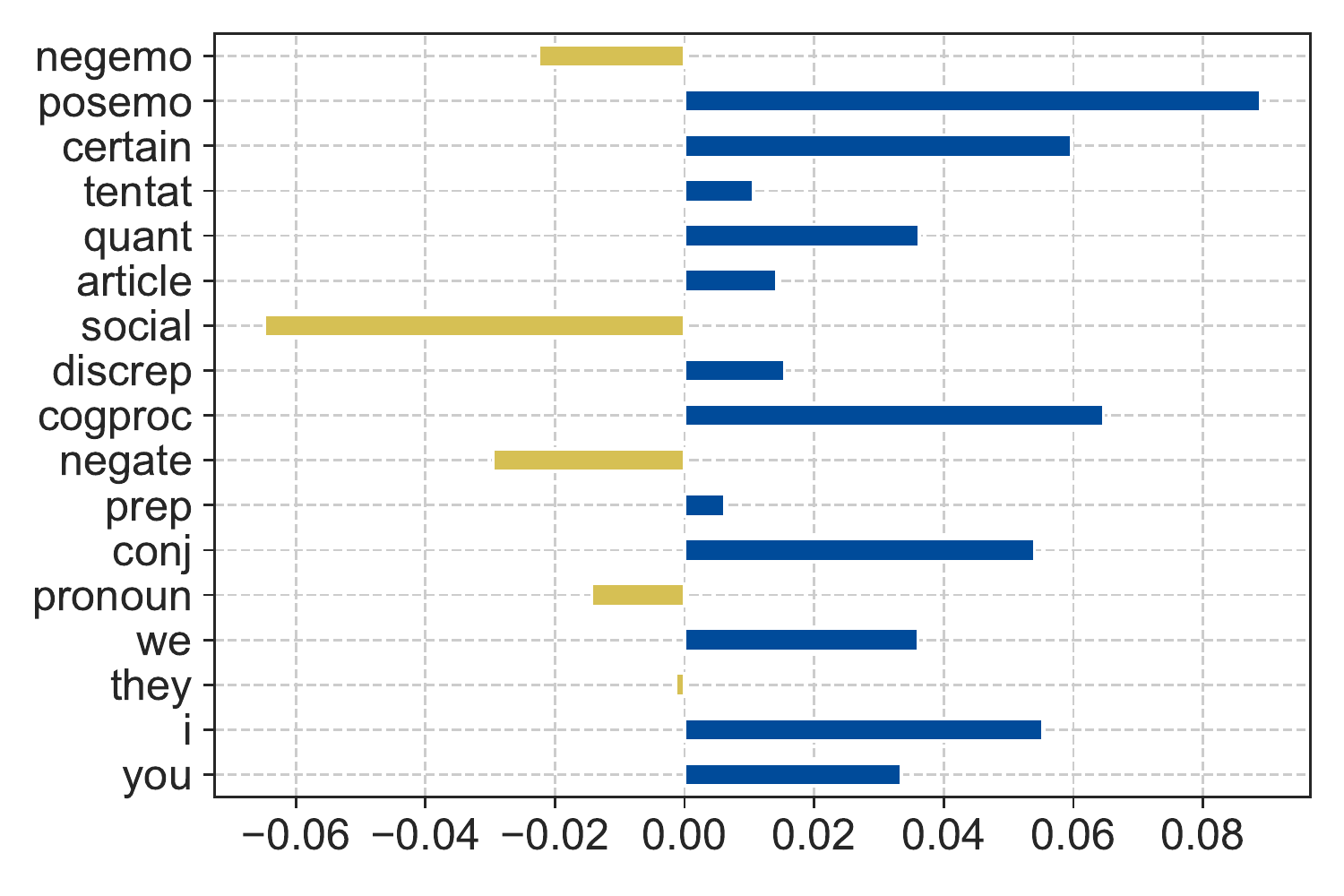}
            \caption[]%
            {{ Bot TockTock Alignment}}    
            \label{fig:bot-ticktock}
        \end{subfigure}
        \vskip\baselineskip
        \begin{subfigure}[b]{0.475\textwidth}   
            \centering 
            \includegraphics[width=\textwidth]{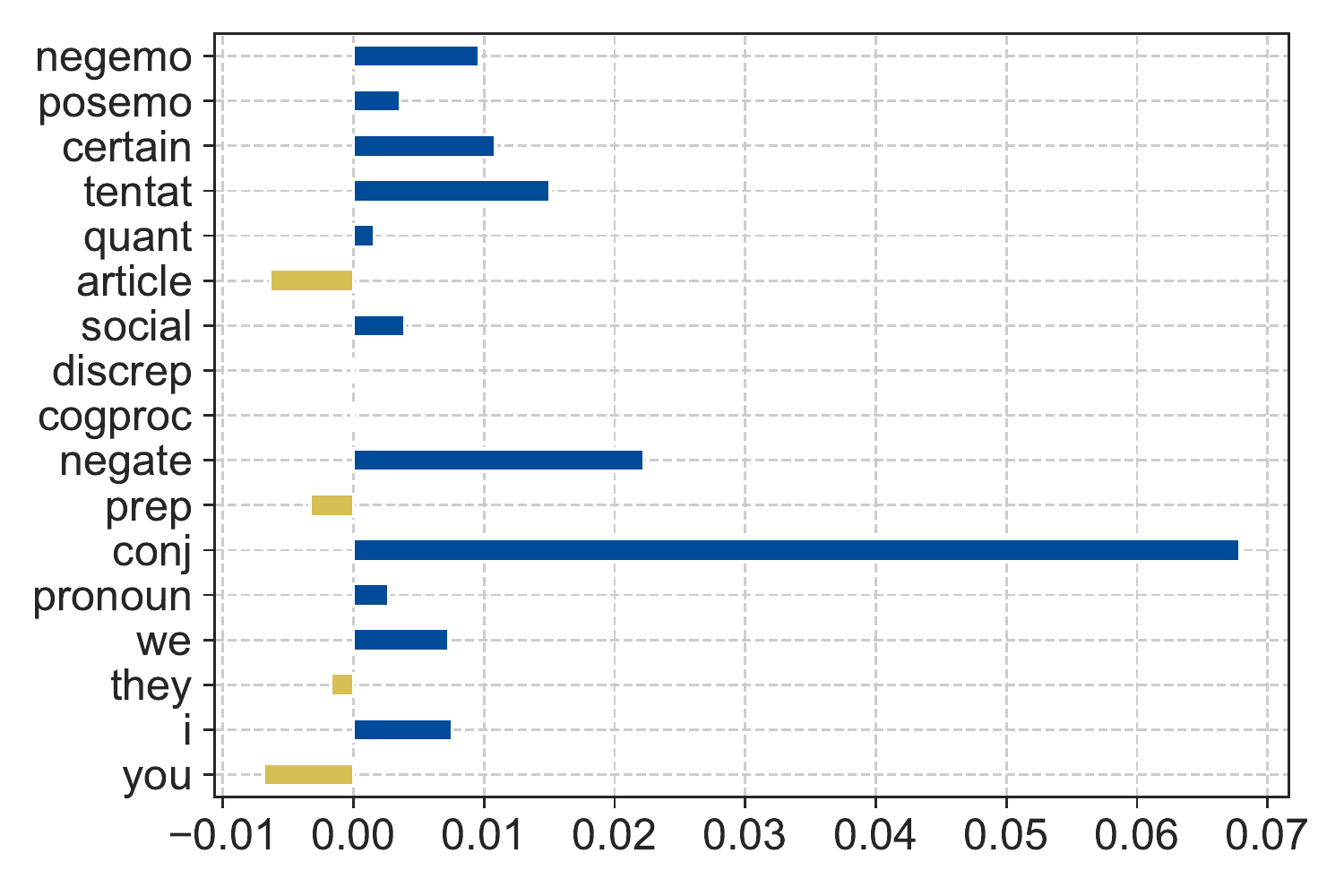}
            \caption[]%
            {{ Human INTERMEDIATE Alignment}}    
            \label{fig:human-inter}
        \end{subfigure}
        \hfill
        \begin{subfigure}[b]{0.475\textwidth}   
            \centering 
            \includegraphics[width=\textwidth]{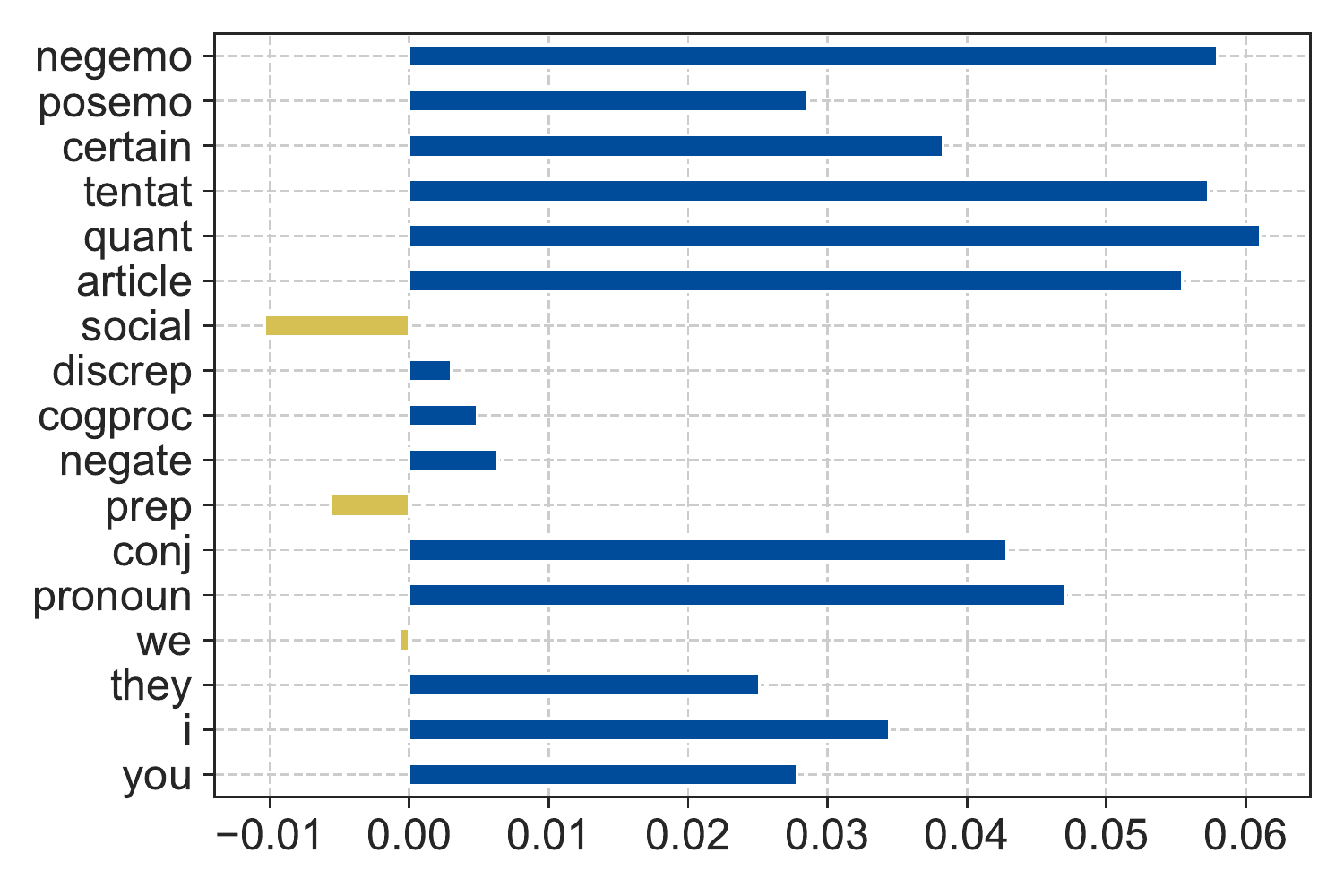}
            \caption[]%
            {{ Bot INTERMEDIATE Alignment}}    
            \label{fig:bot-inter}
        \end{subfigure}
        \caption[]
        { Alignment scores for the TickTock and INTERMEDIATE datasets.\vspace{-1em}} 
        \label{fig:convAIalign-and-base}
    \end{figure*}

In Table~\ref{tab:features}, we analyze the Random Forest's importance scores for the best model (trained on dataset P's human responses) features using the model that combines human LIWC and accommodation features. The most informative feature is pronoun accommodation. See Section~\ref{sec:discussions} for more details. Other informative features include the human's use of social, quant (quantitative), and tentative words.

\subsection{Accommodation Analysis}\label{sec:results:analysis}

In this section, we analyze the group-level (averaged) accommodation scores for the INTERMEDIATE and TickTock datasets.  Again, note that the group-level scores are dataset averages for each group (human and bot), the conversation-level estimates can vary from the group scores. The accommodation analysis is displayed in Figure~\ref{fig:convAIalign-and-base}. The  bot results (Figures~\ref{fig:bot-inter} and~\ref{fig:bot-ticktock}) indicate how the bot responds to a human, and the human scores (Figures~\ref{fig:human-inter} and~\ref{fig:human-ticktock}) represent how a human responds to a bot. We report all LIWC categories, including Pronoun usage (e.g., pronoun, I, you, we, and they) and Rhetorical words (e.g., social, cogproc, posemo, and negemo) results. Find the analysis for all of the datasets in the Supplementary Material.

Overall, some indicators are robust across datasets, whereas others can vary. For instance, for the general pronoun category, we find that human generally accommodate less towards bots than bots accommodate towards them. The pronoun accommodation pattern for pronouns is robust across both datasets. Similarly, for both datasets, humans accomodate more often than bots for social words. Yet, for negative emotion words (negemo), we find that Bot's accommodate less than humans in the TickTock dataset, yet they accommodate more in the INTERMEDIATE dataset. In contrast, in the ConvAI2 and DailyDialog datasets, human accommodation is the same \textit{on average} when two humans participate in a conversation together (not shown in the Figures).

\section{Discussion}\label{sec:discussions}

Why is humans' accommodation such a powerful predictor that they are in a conversation with a bot---even when they are supposedly treating the conversation the same as a human-human interaction? Overall, we found two major reasons to explain the model's high performance. First, modern bots from the transformer to retrieval-based models generate bad responses. For instance, in the conversation below from the VOLUNTEER dataset, the human and bot have already had a conversation about hair color:
\begin{Verbatim}[fontsize=\small]
Bot:   What is your hair color?
Human: I said its black....
Bot:   I know the feeling.
Human: What?
\end{Verbatim}
From the example, we can see that the bot is asking about hair color, which was already discussed, then the bot responds with a nonsensical output, ``I know the feeling.'' Thus, we see that the final human response does not accommodate with with``I'' nor any first singular pronoun. The lower alignment score for humans (compared to bots) for the word category I in Figure~\ref{fig:convAIalign-and-base} supports this finding. Furthermore, from Table~\ref{tab:features}, we know that accommodation in terms of pronoun usage is highly predictive of a human-bot conversation. Similar posts can be found in all datasets where bad outputs can explain why humans align differently concerning ``I'' in bot-human interactions than human-human interactions. This finding is interesting because it is similar to the findings in human-human interactions where alignment is positively correlated with understanding in cooperative decision-making tasks~\cite{fusaroli2012coming,kacewicz2014pronoun}. If humans don't understand the conversation, they align less, at least with regards to pronoun usage in our case. The next example from the VOLUNTEER dataset is a better, more coherent conversation:
\begin{Verbatim}[fontsize=\small]
Human: What is your Hobby?
Bot:   I love music! I love Taylor Swift
Human: Me too. And what about Iggy Pop?
\end{Verbatim}
We can see that the final human response aligns with the first person singular word ``me'' to the bot's use of ``I''. A potential use that warrants future work in the area of bots evaluation involves using accommodation information as an additional evaluation metric in human evaluation studies if certain types of accommodation correlate with bad system responses. If human ratings of bot performance are noisy, the use of accomodation information can provide insight into how well a bot performs, if a human interacts with it differently than they do in human-human conversations.

The second reason for the significant predictive performance can be seen in adversarial conversations between the human and bot. Specifically, in instances where the human is trying to trick the bot or simply trying to ask questions, bot-human alignment can differ from human-human conversations. The example below from the TickTock~dataset
\begin{Verbatim}[fontsize=\small]
Human: Are you a robot?
Bot:   No.
Human: Where are you from?
\end{Verbatim}
shows repeated ``\textit{You}'' usage by the user, repeatedly asking multiple questions. Again, differences in pronoun usage can reflect adversarial users. Again, in terms of bot evaluation, if a human is acting adversarially, and it is possible to differentiate adversarial interactions from the human not understanding, this can provide insight into .
More examples can be found in the Supplementary~Material.

\section{Conclusion}
This study has is an exploration of the levels of linguistic alignment in human-bot conversations and presents evidence about how bot-human interactions are different than human-human conversations. This work's insights have implications for future bot detection methodologies. How humans and bots come together and converse manifested by their level of agreement and disagreement is a first step towards understanding human interactions with general conversational~agents. Furthermore, as future work, we will explore the use of linguistic alignment to improve bot evaluation frameworks involving human-bot interactions. Moreover, we will also explore methods of controlling the accommodation level of bots themselves. Given the implicit social implications of certain types of accommodation, controlling it can help important tasks (e.g., counseling or customer service-related bots).


\bibliography{acl2020}
\bibliographystyle{acl_natbib}

\appendix

\section{Expanded Alignment Analysis}

\begin{figure*}[t]
        \centering
        \begin{subfigure}[b]{0.475\textwidth}
            \centering
            \includegraphics[width=\textwidth]{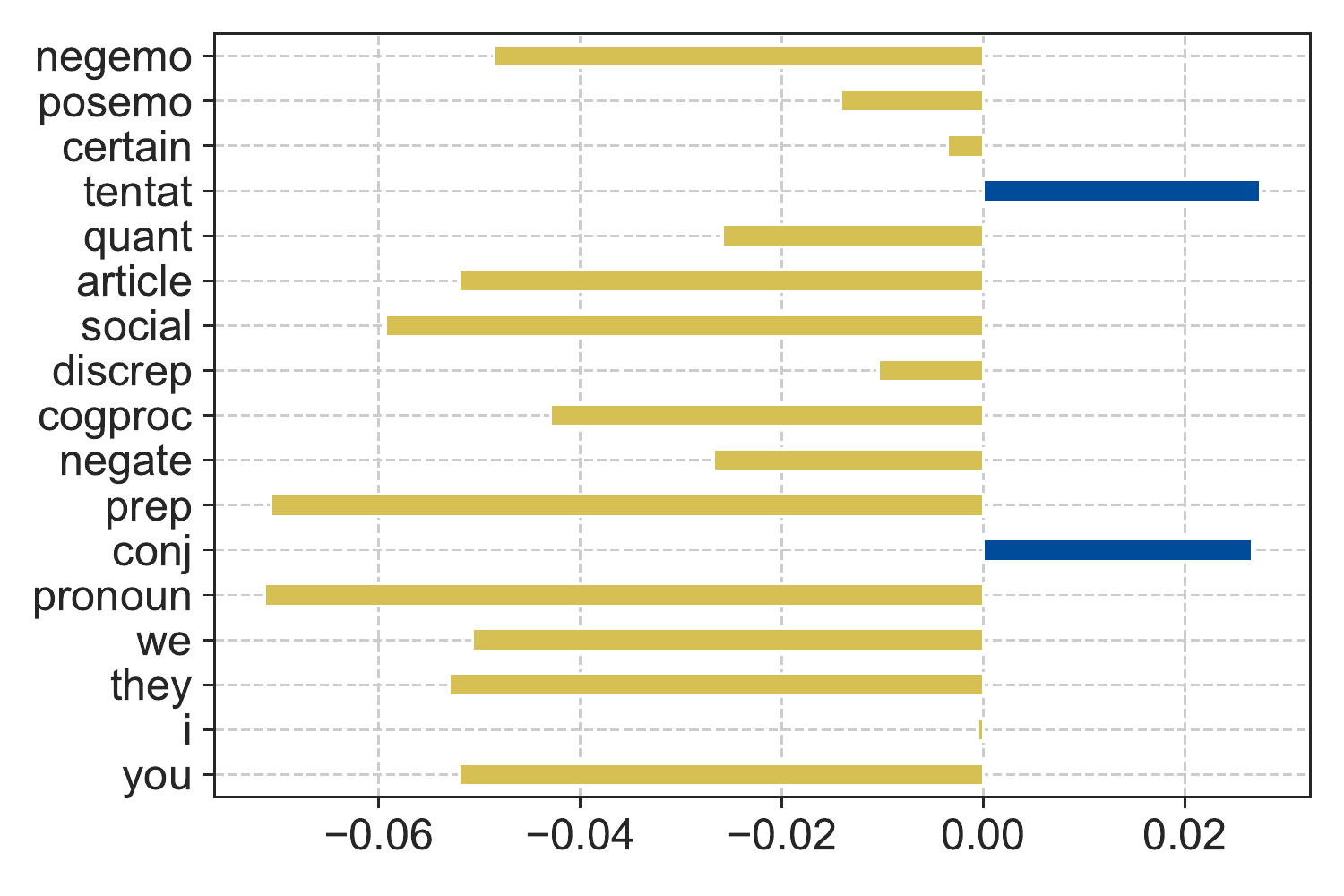}
            \caption[Network2]%
            {{ Human IRIS Alignment}}    
            \label{fig:11}
        \end{subfigure}
        \hfill
        \begin{subfigure}[b]{0.475\textwidth}  
            \centering 
            \includegraphics[width=\textwidth]{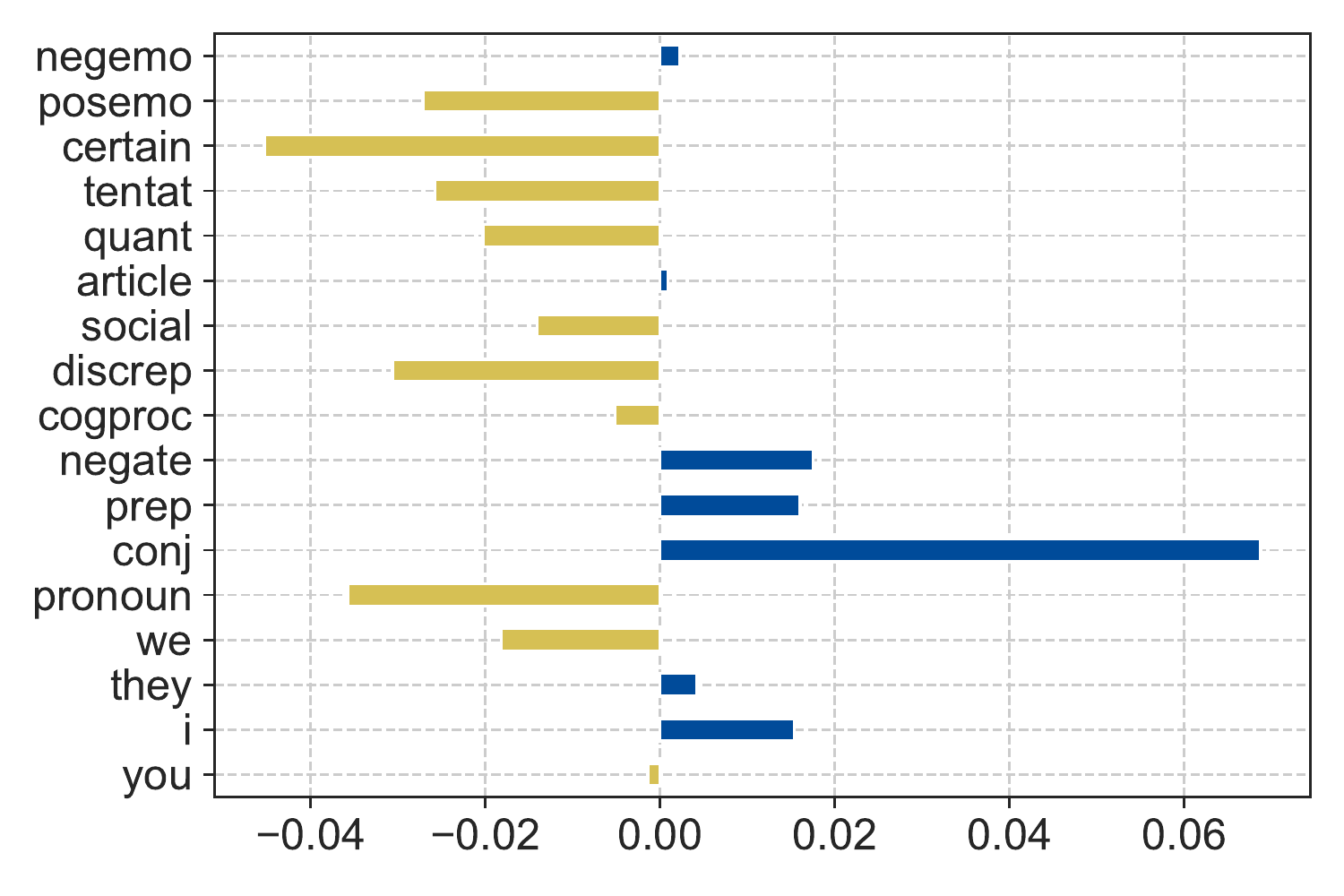}
            \caption[]%
            {{ Bot IRIS Alignment}}    
            \label{fig:22}
        \end{subfigure}
        \vskip\baselineskip \vspace{-1em}
        \begin{subfigure}[b]{0.475\textwidth}   
            \centering 
            \includegraphics[width=\textwidth]{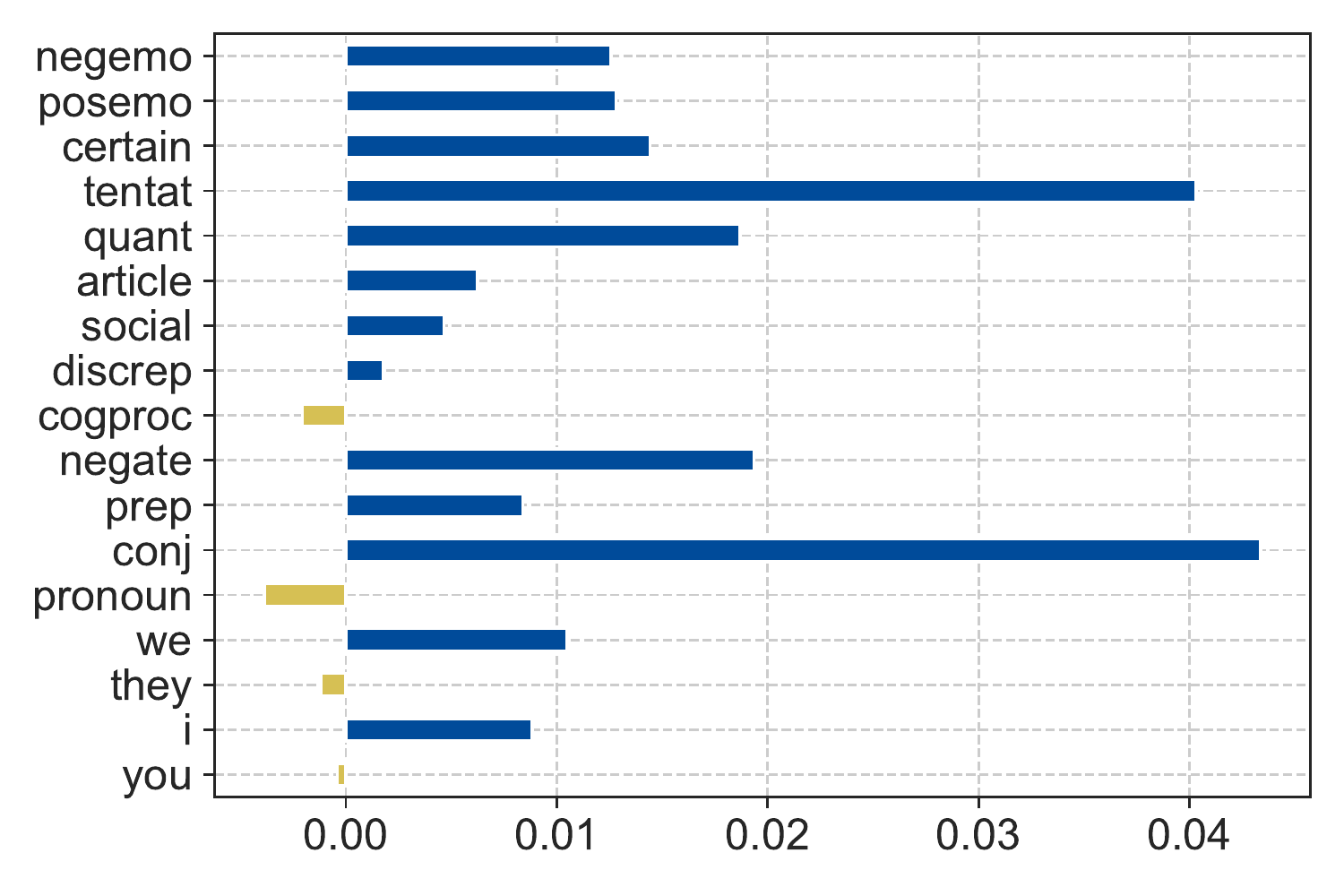}
            \caption[]%
            {{ Human Tolokers Alignment}}    
            \label{fig:33}
        \end{subfigure}
        \hfill
        \begin{subfigure}[b]{0.475\textwidth}   
            \centering 
            \includegraphics[width=\textwidth]{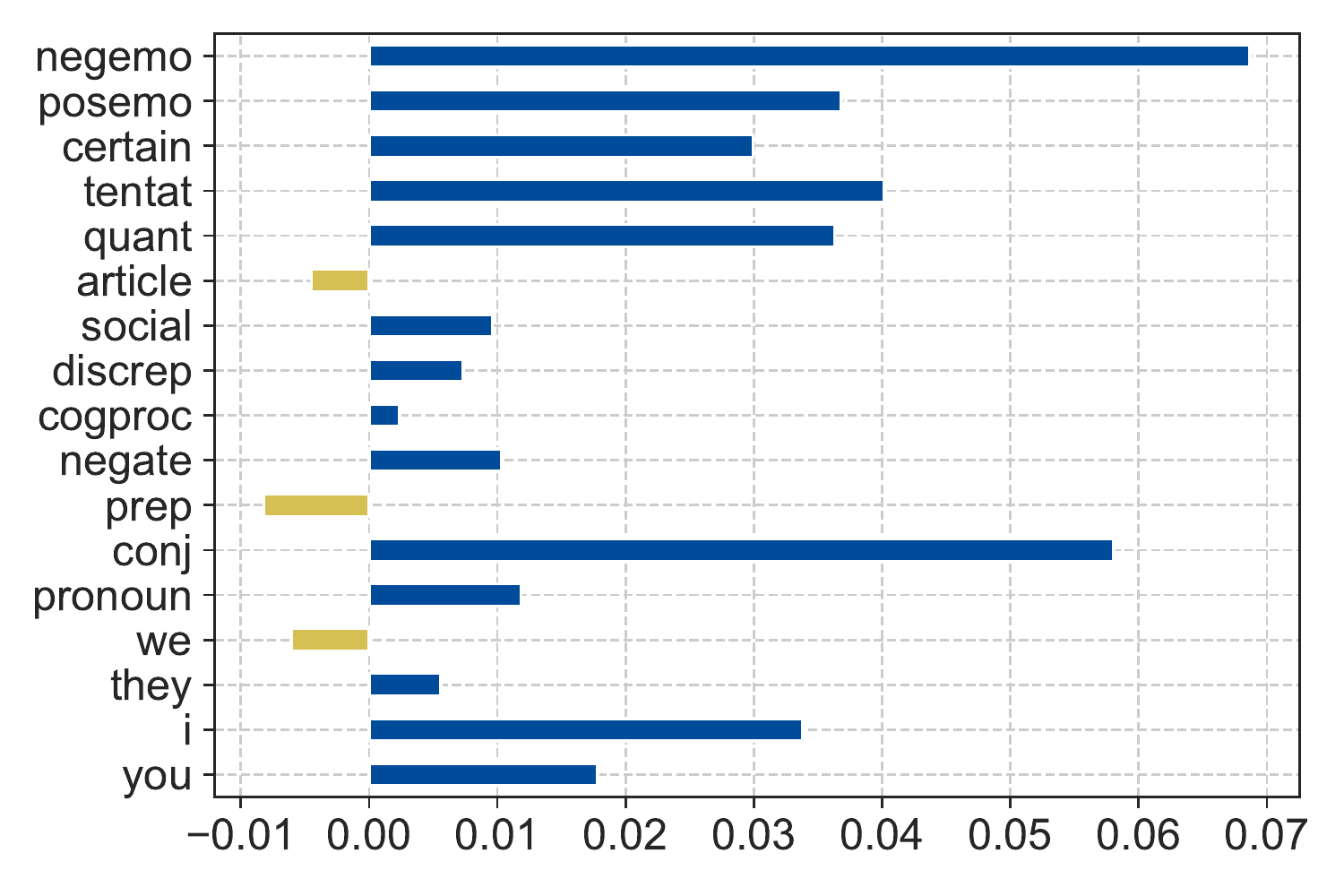}
            \caption[]%
            {{ Bot Tolokers Alignment}}    
            \label{fig:44}
        \end{subfigure}
        \caption[]
        {Alignment scores for the IRIS and Tolokers dataset} 
        \label{fig:55}
    \end{figure*}

The complete group-level alignment scores for the IRIS and Tolokers datasets can be found in Figures~\ref{fig:11},~\ref{fig:22},~\ref{fig:33}, and \ref{fig:44}.

\section{Example Conversations}


Here we present examples of conversations between humans and bots that are representative of good conversational structure in terms of alignment between the two parties such as the usage of pronoun ``\textit{I}'' in the first example.

\begin{Verbatim}[fontsize=\small]
Bot:   I love iphone!
       i just bought new iphone!
Human: Thats good for you  i m not
       very into new tech
\end{Verbatim}
\vspace{1em}

\begin{Verbatim}[fontsize=\small]
Human: I am fine. I like hip hop.
       What about you?
Bot:   I like r b and pop
       are you 100 years old?
Human: What? No. You are rude
\end{Verbatim}
 \vspace{1em}
 
\begin{Verbatim}[fontsize=\small]
Human: Do you have dogs?
Bot:   Yes  i have two dogs  and a cat.
Human: Cool  I love dogs.
       But i have siblings instead...
Bot:   I have two brothers and
       a sister  and a sister.
\end{Verbatim}
 \vspace{1em}

Here conversations between humans and bots involve irregular pronoun usage that are representative of a break in the conversational pattern and little or no alignment between the two parties such as the usage of pronouns ``\textit{You}'' and resultant use of ``\textit{I}'' reply  in the first example.

\begin{Verbatim}[fontsize=\small]
Human: See you then!
Bot:   I m not going to see you.
\end{Verbatim}
 \vspace{1em}

\begin{Verbatim}[fontsize=\small]
Bot:   I am 24 and I have a dog named
       her name is named name
Human: Really her name is name?
Bot:   She is a good name.
\end{Verbatim}
 \vspace{1em}

\begin{Verbatim}[fontsize=\small]
Human: who is obama
Bot:   i partially have interest in that
Human: can you tell me more about obama
Bot:   i hate pigs i live a farm
\end{Verbatim}
 \vspace{1em}

These conversations signify a complete breakdown of conversational structure with respect to alignment, as there is none. The chats seem haphazard and vague with either the bot or the human replying in an ambiguous manner.

\begin{Verbatim}[fontsize=\small]
User:   Good afternoon
System: Well  you know  I'm not sure.
        What do you think about movies
User:   That didn't make any sense
\end{Verbatim}
 \vspace{1em}

\begin{Verbatim}[fontsize=\small]
User:   hi again	
System: Of course. Any thoughts?
User:   no
\end{Verbatim}
 \vspace{1em}

\begin{Verbatim}[fontsize=\small]
User:   Hello	
System: Do not argue with an idiot.
        He will drag you down to his
        level and beat you with
        experience.
User:   Um  Okay  that was random
\end{Verbatim}

\end{document}